\documentclass[conference]{IEEEtran}
\IEEEoverridecommandlockouts

\usepackage{cite}
\usepackage{amsmath,amssymb,amsfonts}
\usepackage{algorithmic}
\usepackage{graphicx}
\usepackage{textcomp}
\usepackage{enumitem}
\usepackage{booktabs}    % nicer rules
\usepackage{multirow}
\usepackage{array}
\usepackage{tikz}
\usepackage{caption}
\usepackage[table]{xcolor}
\usepackage{colortbl}
\usepackage{hyperref}

\definecolor{softgreen1}{RGB}{240,250,240} % very light
\definecolor{softgreen2}{RGB}{225,245,225}
\definecolor{softgreen3}{RGB}{205,238,205} % light-mid
\definecolor{softred}{RGB}{245,220,220}
\definecolor{softblue}{RGB}{220,230,245}
\definecolor{softwhite}{RGB}{255,255,255}

\def\BibTeX{{\rm B\kern-.05em{\sc i\kern-.025em b}\kern-.08em
    T\kern-.1667em\lower.7ex\hbox{E}\kern-.125emX}}

\begin{document}

\title{ $\mathcal{S}$tride-Net: Fairness-Aware Disentangled Representation Learning for Chest X-Ray Diagnosis}

\author{\IEEEauthorblockN{Darakshan Rashid$^1$, Raza Imam$^2$, Dwarikanath Mahapatra$^3$, Brejesh Lall$^1$}
\IEEEauthorblockA{$^1$Indian Institute of Technology Delhi, $^2$MBZUAI, $^3$Khalifa University\\
bsz228540@iitd.ac.in, raza.imam@mbzuai.ac.ae, dwarikanath.mahapatra@ku.ac.ae, brejesh.lall@ee.iitd.ac.in}
}

\maketitle
\begin{abstract}
Deep neural networks for chest X-ray classification achieve strong average performance, yet often underperform for specific demographic subgroups, raising critical concerns about clinical safety and equity. 
Existing debiasing methods frequently yield inconsistent improvements across datasets or attain fairness by degrading overall diagnostic utility, treating fairness as a post hoc constraint rather than a property of the learned representation.
In this work, we propose $\mathcal{S}$tride-Net ($\mathcal{S}$ensitive a$\mathcal{T}$tribute $\mathcal{R}$esilient learning v$\mathcal{I}$a $\mathcal{D}$isentanglement and learnable masking with $\mathcal{E}$mbedding alignment), a fairness-aware framework that learns disease-discriminative yet demographically invariant representations for chest X-ray analysis. $\mathcal{S}$tride-Net operates at the patch level, using a learnable stride-based mask to select label-aligned image regions while suppressing sensitive attribute information through adversarial confusion loss. To anchor representations in clinical semantics and discourage shortcut learning, we further enforce semantic alignment between image features and BioBERT-based disease label embeddings via Group-Optimal Transport.
We evaluate $\mathcal{S}$tride-Net on the MIMIC-CXR and CheXpert benchmarks across race and intersectional race–gender subgroups. Across architectures including ResNet and Vision Transformers, $\mathcal{S}$tride-Net consistently improves fairness metrics while matching or exceeding baseline accuracy, achieving a more favorable accuracy–fairness trade-off than prior debiasing approaches. 
% Our code will be made publicly available upon paper acceptance.
Our code is available \href{https://github.com/Daraksh/Fairness_StrideNet}{Here}.
\end{abstract}


% Sections

\section{Introduction}

% Artificial intelligence (AI) has made rapid advances in medical imaging, driven by the availability of large-scale datasets and increasingly expressive deep learning architectures~\cite{irvin2019chexpert, mienye2025deep}. 
In chest X-ray (CXR) analysis, deep models now approach or exceed human-level performance on several diagnostic tasks~\cite{irvin2019chexpert, mienye2025deep} and are increasingly considered for clinical deployment in screening, triage, and decision support~\cite{esteva2021deep, liawrungrueang2025artificial}. Despite these advances, growing evidence shows that models with strong average performance can exhibit systematic disparities across demographic subgroups, such as race and gender~\cite{obermeyer2019dissecting, seyyed2020chexclusion}. These disparities are not merely statistical artifacts; they translate into real clinical risks, particularly for underrepresented and intersectional populations.

A striking example arises in the prediction of the clinically critical \emph{``No Finding''} label. This label functions as a gatekeeper in clinical workflows, signaling the absence of detectable pathology and often influencing downstream diagnostic decisions, follow-up imaging, and allocation of care. Errors in ``No Finding'' prediction are therefore asymmetric in their consequences: false negatives may delay diagnosis, while false positives can trigger unnecessary interventions. When such errors disproportionately affect specific demographic groups, they can exacerbate existing healthcare inequities.

\vspace{0.2cm}

\textbf{Fairness challenges in chest X-ray modeling}  
Recent studies have shown that fairness interventions in CXR models often yield inconsistent or brittle improvements across datasets, sensitive attributes, and deployment settings~\cite{zong2022medfair}. A core difficulty lies in the fact that demographic information can be subtly encoded in medical images through acquisition protocols, anatomical variation, or dataset-specific biases. Standard empirical risk minimization (ERM) objectives readily exploit such correlations if they improve predictive accuracy on the training distribution, even when those correlations are clinically irrelevant or harmful.

As a result, CXR models may implicitly rely on sensitive attributes as shortcuts for disease prediction. This phenomenon is particularly problematic for decision-critical labels such as ``No Finding'', where small representational biases can have outsized downstream effects. Moreover, these issues are often amplified under distribution shift, when models are deployed in populations that differ from the training cohort~\cite{mehrabi2021survey, zong2022medfair}. Achieving equitable performance therefore requires representations that are highly predictive of pathology while being invariant to sensitive attributes; a goal that is fundamentally challenging under standard optimization.

\vspace{0.2cm}

\begin{figure}[t]
    \centering
    \begin{tikzpicture}
        \node[anchor=south west, inner sep=0] (img) at (-0.2,0)
        {\includegraphics[width=\linewidth]{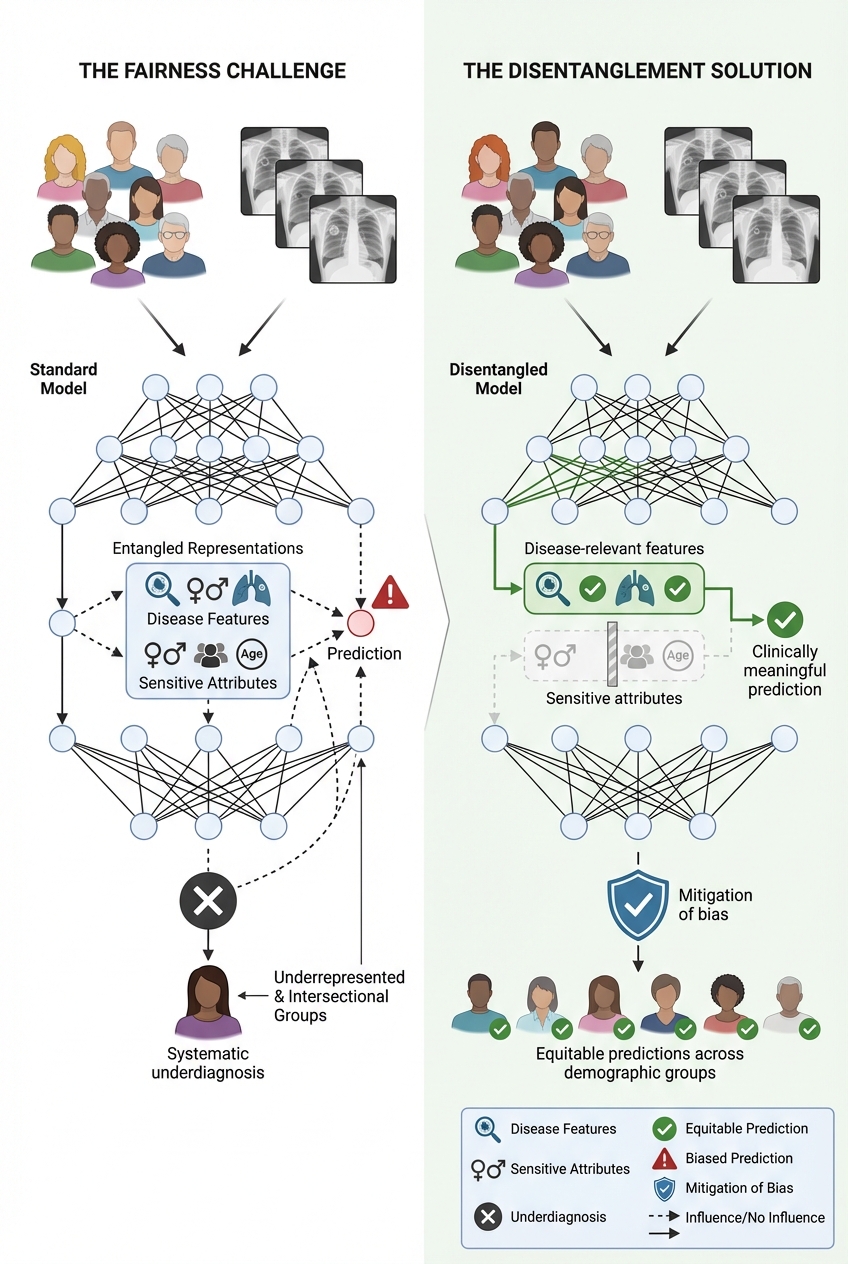}};
        \node[
            anchor=south west,
            text width=0.40\linewidth,
            align=left,
            font=\tiny
        ] at ([xshift=6mm,yshift=1mm]img.south west)
        {Standard models entangle disease-relevant features with sensitive attributes, 
        leading to biased predictions, while disentangled models isolate clinically 
        meaningful features to mitigate bias and promote equitable performance across 
        demographic groups.};
    \end{tikzpicture}
    \caption{Conceptual illustration of fairness and disentanglement in chest X-ray classification.}
    \label{fig:fairness_concept}
    \vspace{-0.4cm}
\end{figure}

\textbf{Limitations of existing fairness approaches}  
Most fairness-aware learning methods can be broadly categorized as pre-processing, in-processing, or post-processing techniques~\cite{mehrabi2021survey}. Group fairness objectives such as demographic parity or equalized odds often impose explicit constraints on predictions, but may induce undesirable trade-offs with accuracy or lead to ``leveling down'', where performance is uniformly degraded across groups to satisfy fairness criteria~\cite{zietlow2022leveling}. In clinical settings, such trade-offs are particularly problematic, as they can reduce overall diagnostic utility.

In-processing approaches, including GroupDRO~\cite{sagawa2019distributionally}, MMD Match~\cite{pfohl2021empirical}, and adversarial debiasing~\cite{wadsworth2018achieving}, aim to mitigate bias during training, yet empirical evaluations show that they do not consistently outperform strong ERM baselines in medical imaging tasks~\cite{zong2022medfair}. Parameter-efficient fine-tuning (PEFT) methods have recently been proposed as a lightweight fairness mechanism~\cite{dutt2023fairtune}, but their effectiveness remains sensitive to hyperparameter choices and computational budgets.

Disentanglement-based methods offer a principled alternative by explicitly factorizing latent representations into independent components~\cite{tartaglione2021end}, \cite{rahman2025decoupling}. However, despite their conceptual appeal, disentanglement techniques have rarely been integrated with fairness objectives in medical imaging, and even more rarely at the level of localized, patch-based representations where clinically meaningful signals reside.

\vspace{0.2cm}

\textbf{Toward fair and semantically grounded representations}  
A promising direction to address these limitations is to intervene directly at the level of representation learning. Rather than constraining outputs post hoc, one can aim to structure the latent space so that disease-relevant features are explicitly isolated from sensitive attribute information. In the context of CXR, this suggests a model that (i) operates at the patch level to localize clinically meaningful regions, (ii) aligns visual features with disease semantics to discourage shortcut learning, and (iii) actively suppresses demographic information through adversarial disentanglement.

\begin{figure*}
\centering
\includegraphics[width=\linewidth]{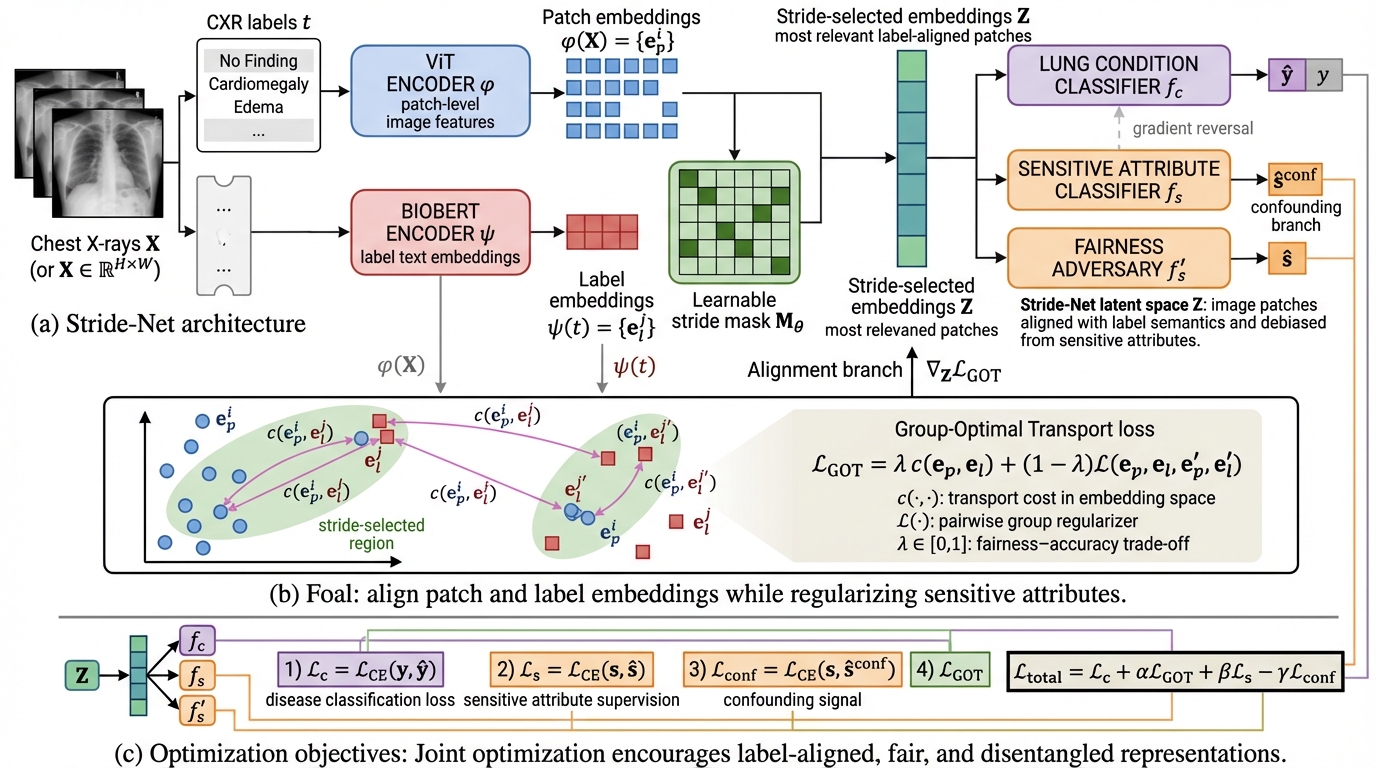}
\caption{$\mathcal{S}$tride-Net architecture for fair chest X-ray classification.
A ViT encoder $\phi$ extracts patch embeddings from an input X-ray, while disease labels are encoded by BioBERT $\psi$. A learnable stride mask selects label-aligned patches to form the latent representation $Z$, which is used for disease prediction. Group-Optimal Transport enforces semantic alignment and fairness, while adversarial sensitive-attribute classifiers promote demographic invariance.}
\label{fig:stride_arch_ieee}
\end{figure*}

Motivated by these principles, we introduce $\mathcal{S}$tride-Net, a fairness-aware representation learning framework for chest X-ray classification. As illustrated in Fig.~\ref{fig:stride_arch_ieee}, $\mathcal{S}$tride-Net employs a learnable stride mask to select label-aligned image patches and route them into a shared latent space. Semantic alignment between visual patches and disease labels is enforced via Group-Optimal Transport (GOT) using BioBERT-based label embeddings, anchoring representations in clinically meaningful semantics. In parallel, adversarial confusion losses discourage the encoding of sensitive attributes, promoting demographic invariance without sacrificing predictive power.

\vspace{0.2cm}

\textbf{Contributions}  
Our contributions are summarized as follows:
\begin{itemize}
\item[--] We propose $\mathcal{S}$tride-Net, a fairness-aware disentanglement framework for chest X-ray classification that explicitly separates disease-relevant features from sensitive demographic information using a learnable stride-based mask.
\item[--] We integrate semantic supervision via Group-Optimal Transport alignment between image patches and BioBERT-based disease label embeddings, grounding visual representations in clinical semantics.
\item[--] We demonstrate consistent improvements in fairness across race and intersectional race–gender subgroups, with a particular focus on underdiagnosis in the clinically important ``No Finding'' task.
\item[--] We validate our approach on MIMIC-CXR and CheXpert, showing that $\mathcal{S}$tride-Net achieves a favorable accuracy–fairness trade-off compared to ERM baselines and established debiasing methods.
\end{itemize}

\section{Proposed Methodology}
\label{model}

The central goal of $\mathcal{S}$tride-Net is to learn disease-discriminative visual representations from chest X-rays that are explicitly aligned with clinical label semantics while being invariant to sensitive demographic attributes such as race or gender. As illustrated in Fig.~\ref{fig:stride_arch_ieee}, the proposed architecture integrates patch-level visual encoding, label-aware semantic alignment, and adversarial fairness regularization within a unified, end-to-end trainable framework. By operating at the level of localized image patches and explicitly enforcing alignment with disease label embeddings, $\mathcal{S}$tride-Net mitigates shortcut learning driven by demographic confounders and promotes fair, interpretable decision-making.
\vspace{0.2cm}

\textbf{Problem Formulation}
Let $X \subset \mathbb{R}^{H \times W}$ denote the space of chest X-ray images and $Y$ the space of disease labels. We consider a supervised learning setting with a dataset of $N$ samples
\[
\mathcal{D} = \{(x_i, y_i, s_i)\}_{i=1}^{N},
\]
where $x_i \in X$ is a chest X-ray image, $y_i \in Y$ is the corresponding disease label, and $s_i \in S$ denotes a sensitive attribute such as race or gender. The sensitive attribute is assumed to be available during training but not at inference time.

The objective of $\mathcal{S}$tride-Net is threefold:  
(i) to accurately predict disease labels $\hat{y}$ from chest X-rays,  
(ii) to minimize the information about sensitive attributes $s$ encoded in the learned latent representation, and  
(iii) to ensure that disease predictions are grounded in clinically relevant image regions rather than spurious demographic correlations.

Unlike post-hoc debiasing approaches, $\mathcal{S}$tride-Net enforces fairness directly during representation learning by structuring the latent space to be label-aligned and demographically invariant.
\vspace{0.2cm}

\textbf{Patch-Level Visual Encoding}
Given an input chest X-ray image $x \in X$, we first extract localized visual features using a Vision Transformer (ViT) encoder $\phi$. The image is divided into a grid of non-overlapping patches, and each patch is mapped to a patch embedding through the encoder:
\[
\phi(x) = \{ e_p^i \}_{i=1}^{P},
\]
where $e_p^i \in \mathbb{R}^d$ denotes the embedding of the $i$-th image patch, $P$ is the total number of patches, and $d$ is the embedding dimension.

Patch-level representations preserve spatial granularity, enabling the model to selectively attend to localized anatomical regions such as lung fields or cardiac silhouettes. This design is crucial for mitigating bias, as demographic signals often manifest globally, whereas disease evidence is typically localized.
\vspace{0.2cm}

\textbf{Label Semantic Encoding}
To inject clinical semantics into the learning process, we encode disease labels using a pretrained BioBERT encoder $\psi$. Each disease label $t \in Y$ is mapped to a semantic embedding:
\[
\psi(t) = \{ e_l^j \}_{j=1}^{L},
\]
where $e_l^j \in \mathbb{R}^d$ represents the embedding of the $j$-th disease label token, and $L$ is the number of labels.

By embedding labels in the same latent space as image patches, $\mathcal{S}$tride-Net enables direct semantic alignment between visual evidence and disease concepts. This alignment acts as a form of weak supervision that encourages the model to rely on medically meaningful image regions.
\vspace{0.2cm}

\textbf{Stride-Based Patch Selection}
Not all image patches contribute equally to disease prediction. To identify the most label-relevant visual regions, we introduce a learnable stride mask $M_\theta \in \mathbb{R}^{P \times L}$, parameterized by $\theta$. This mask computes relevance scores between patch embeddings $\{e_p^i\}$ and label embeddings $\{e_l^j\}$.

The stride mask selectively activates a subset of patches that are most semantically aligned with the disease labels. Formally, the stride-selected latent representation is given by:
\[
Z = \mathrm{StrideSelect}(\phi(x), \psi(t), M_\theta),
\]
where $Z \subset \{e_p^i\}$ consists of the most relevant, label-aligned patch embeddings.

This stride-based mechanism enforces sparsity and interpretability, ensuring that downstream predictions are driven by clinically relevant regions rather than background artifacts or demographic proxies.
\vspace{0.2cm}

\textbf{Group-Optimal Transport Alignment}
To further enforce semantic consistency and fairness, $\mathcal{S}$tride-Net employs a Group-Optimal Transport (GOT) loss to align patch embeddings with label embeddings while regularizing sensitive attribute disparities. The GOT loss is defined as:
\[
\mathcal{L}_{\mathrm{GOT}} = \lambda \, c(e_p, e_l) + (1 - \lambda) \, \mathcal{L}(e_p, e_l, e_p', e_l'),
\]
where $c(\cdot, \cdot)$ denotes the transport cost in the embedding space, $\mathcal{L}(\cdot)$ is a pairwise group regularizer, and $\lambda \in [0,1]$ controls the trade-off between accuracy and fairness.

Intuitively, the first term encourages patch embeddings to align closely with their corresponding disease label embeddings, while the second term enforces group-wise consistency across different sensitive attribute values. This alignment reduces representational discrepancies that could otherwise lead to biased predictions.
\vspace{0.2cm}

\textbf{Disease Classification Branch}
The stride-selected latent representation $Z$ is fed into a disease classifier $f_c$ to predict disease labels:
\[
\hat{y} = f_c(Z).
\]
The disease classification loss is defined using standard cross-entropy:
\[
\mathcal{L}_c = \mathcal{L}_{\mathrm{CE}}(y, \hat{y}).
\]
Because $Z$ contains only label-aligned, stride-selected patches, the classifier is forced to rely on clinically meaningful evidence rather than spurious correlations.
\vspace{0.2cm}

\textbf{Sensitive Attribute Prediction and Adversarial Fairness}
To explicitly remove sensitive information from the latent space, we introduce two sensitive attribute classifiers. The first classifier $f_s$ predicts the sensitive attribute from $Z$:
\[
\hat{s} = f_s(Z),
\]
with supervision loss
\[
\mathcal{L}_s = \mathcal{L}_{\mathrm{CE}}(s, \hat{s}).
\]

In parallel, a fairness adversary $f_s'$ is trained using gradient reversal to maximize confusion with respect to sensitive attribute prediction:
\[
\hat{s}^{\mathrm{conf}} = f_s'(Z),
\]
with corresponding loss
\[
\mathcal{L}_{\mathrm{conf}} = \mathcal{L}_{\mathrm{CE}}(s, \hat{s}^{\mathrm{conf}}).
\]

The gradient reversal mechanism ensures that the feature extractor learns representations that are predictive of disease while being uninformative of sensitive attributes.
\vspace{0.2cm}

\textbf{Joint Optimization Objective}
All components of $\mathcal{S}$tride-Net are trained jointly in an end-to-end manner. The total loss function is given by:
\[
\mathcal{L}_{\mathrm{total}} =
\mathcal{L}_c
+ \alpha \mathcal{L}_{\mathrm{GOT}}
+ \beta \mathcal{L}_s
- \gamma \mathcal{L}_{\mathrm{conf}},
\]
where $\alpha$, $\beta$, and $\gamma$ are scalar hyperparameters controlling the relative contributions of semantic alignment, sensitive attribute supervision, and adversarial debiasing.

This joint objective encourages the model to learn latent representations that are simultaneously discriminative, label-aligned, and fair.
% \vspace{0.2cm}

% {\textbf{Summary}}

% In summary, $\mathcal{S}$tride-Net integrates patch-level vision encoding, label-semantic alignment, stride-based patch selection, optimal transport regularization, and adversarial fairness learning into a coherent architecture. By enforcing fairness at the level of representation learning, $\mathcal{S}$tride-Net provides a principled and effective framework for fair chest X-ray classification.
\section{Experimental Setup}

\textbf{Datasets and Preprocessing}
We evaluate $\mathcal{S}$tride-Net on two widely used large-scale chest X-ray benchmarks: MIMIC-CXR~\cite{johnson2019mimic} and CheXpert~\cite{irvin2019chexpert}. Following prior fairness-focused work in medical imaging~\cite{seyyed2021underdiagnosis}, we consider the binary \emph{No Finding} classification task and analyze performance disparities across sensitive demographic attributes, specifically race and gender.

Sensitive attribute annotations are defined using the same protocol as~\cite{seyyed2021underdiagnosis}. To ensure reliable subgroup analysis, samples with missing sensitive attribute information are excluded. This filtering step is critical for fairness evaluation, as inaccurate subgroup membership can obscure or distort disparity measurements.

All chest X-ray images are resized to $224 \times 224 \times 3$ to match the input resolution of the Vision Transformer backbone. Labels annotated as ``Not mentioned’’ are treated as ``uncertain’’ and assigned a value of $-1$, consistent with standard preprocessing practices. We split each dataset into training and testing sets using an 80/20 stratified split that preserves the distribution of sensitive subgroups. During training, standard data augmentations; including random horizontal flips and small-angle rotations; are applied to improve robustness and generalization \cite{imam2025robustness}.
\vspace{0.2cm}

\textbf{Training Configuration}
All models are trained using the Adam optimizer with a learning rate of $1 \times 10^{-4}$ and a batch size of 64. Training is performed for 20 epochs on a single NVIDIA RTX A6000 GPU. Unless otherwise specified, hyperparameters are held constant across datasets and model variants to ensure fair comparison.

Semantic supervision for the Group-Optimal Transport (GOT) module is provided by fixed label embeddings extracted using BioBERT~\cite{lee2020biobert}. These embeddings encode clinical semantics of disease labels and remain frozen throughout training, allowing the model to align visual features with stable semantic anchors. For Vision Transformer-based variants, we fine-tune a ViT-B model pre-trained on ImageNet
% ~\cite{dosovitskiy2020image}
.
\vspace{0.2cm}

\textbf{Fairness Evaluation Metrics}
We evaluate fairness using two complementary metrics that capture different aspects of subgroup performance equity.

\textit{1) Predictive Quality Disparity (PQD).}  
Let $R$ denote the set of sensitive subgroups, and let $\text{acc}_j$ be the classification accuracy for subgroup $j \in R$. PQD measures the uniformity of predictive performance across subgroups:
\begin{equation}
PQD = \frac{\min_{j \in R} \text{acc}_j}{\max_{j \in R} \text{acc}_j}.
\end{equation}
Higher PQD values indicate smaller performance gaps between the best- and worst-performing subgroups.
% \vspace{0.2cm}

\textit{2) Equality of Opportunity Measure (EOM).}  
Let $M$ denote the number of disease classes, and let $p(\hat{y}=i \mid y=i, r=j)$ be the true positive rate for class $i$ within subgroup $j$. EOM evaluates whether true positive rates are consistent across subgroups:
\begin{equation}
EOM = \frac{1}{M} \sum_{i=1}^{M}
\frac{\min_{j \in R} p(\hat{y}=i \mid y=i, r=j)}{
\max_{j \in R} p(\hat{y}=i \mid y=i, r=j)}.
\end{equation}
Both metrics lie in $[0,1]$, with higher values indicating more equitable model behavior.
\vspace{0.2cm}
\begin{table}[t]
\centering
\scriptsize
\setlength{\tabcolsep}{5pt}
\renewcommand{\arraystretch}{1.15}
\caption{{Overall accuracy and fairness comparison across sensitive attributes}, evaluated across race and intersectional race-gender subgroups. 
% $\mathcal{S}$tride-Net consistently achieves the best balance between predictive performance and fairness, outperforming prior debiasing methods while maintaining high clinical utility.
}
\label{tab:combined_fairness_results_ieee}

\begin{tabular}{lccc|ccc}
\toprule
\multirow{2}{*}{Methods $\downarrow$} &
\multicolumn{3}{c|}{MIMIC-CXR} &
\multicolumn{3}{c}{CheXpert} \\
\cmidrule(lr){2-4}\cmidrule(lr){5-7}
& Acc/Avg & PQD & EOM & Acc/Avg & PQD & EOM \\
\midrule

\multicolumn{7}{l}{\textcolor{gray}{Individual Subgroup: ``Race"}} \\
\midrule
ResNet-18 &
0.780 & 0.850 & 0.680 &
0.898 & 0.896 & 0.480 \\

UBAIA~\cite{seyyed2021underdiagnosis} &
0.789 & \cellcolor{softgreen3}{0.935} & 0.830 &
0.895 & \cellcolor{softgreen3}{0.943} & 0.745 \\

CheXclusion~\cite{seyyed2020chexclusion} &
0.777 & 0.850 & 0.839 &
0.887 & 0.903 & 0.700 \\

 $\mathcal{S}$tride-Net (Ours) &
\cellcolor{softgreen3}{0.805} & \cellcolor{softgreen1}0.922 & \cellcolor{softgreen3}{0.870} &
\cellcolor{softgreen3}{0.903} & \cellcolor{softgreen1}0.928 & \cellcolor{softgreen3}{0.803} \\

\midrule
\multicolumn{7}{l}{\textcolor{gray}{Intersection Subgroup: ``Race-Gender"}} \\
\midrule
ResNet-18 &
0.789 & 0.918 & 0.685 &
0.900 & 0.881 & 0.460 \\

UBAIA~\cite{seyyed2021underdiagnosis} &
0.788 & 0.821 & 0.703 &
0.907 & 0.868 & 0.452 \\

 $\mathcal{S}$tride-Net (Ours) &
\cellcolor{softgreen3}{0.806} & \cellcolor{softgreen3}{0.879} & \cellcolor{softgreen3}{0.705} &
\cellcolor{softgreen3}{0.909} & \cellcolor{softgreen3}{0.897} & \cellcolor{softgreen3}{0.466} \\
\bottomrule
\multicolumn{7}{l}{\tiny{High highlight intensity denote high performance}}

\vspace{-0.3cm}
\end{tabular}
\end{table}

\section{Results}

\textbf{Overall Performance Across Sensitive Subgroups}
We compare $\mathcal{S}$tride-Net against three representative baselines:  
(1) a standard ResNet-18 trained with empirical risk minimization (ERM)~\cite{he2016deep},  
(2) UBAIA~\cite{seyyed2021underdiagnosis}, a fairness-aware method designed to reduce underdiagnosis, and  
(3) CheXclusion~\cite{seyyed2020chexclusion}, which removes spurious correlations via exclusion mechanisms.

Table~\ref{tab:combined_fairness_results_ieee} summarizes results across race and intersectional race-gender subgroups. On MIMIC-CXR, $\mathcal{S}$tride-Net improves average accuracy over UBAIA by $+1.6$ percentage points (0.805 vs.\ 0.789) while also increasing EOM by $+4.0$ points. On CheXpert, similar gains are observed, with accuracy improving by $+0.8$ points and EOM by $+5.8$ points.

These improvements highlight a key advantage of $\mathcal{S}$tride-Net: fairness gains are not achieved by flattening model predictions or sacrificing discriminative power. Instead, stride-based patch selection and semantic alignment enable the model to retain strong disease-relevant signals while suppressing demographic shortcuts.
\vspace{0.2cm}

% \begin{figure}[t]
% \centering
% \includegraphics[width=0.45\textwidth]{Figures/CheXpert_results_classic (1).pdf}
% \includegraphics[width=0.45\textwidth]{Figures/MIMIC_results_classic (1).pdf}
% \caption{\textbf{Accuracy-fairness trade-offs on CheXpert and MIMIC-CXR.}
% Comparison of $\mathcal{S}$tride-Net with representative debiasing methods under the Equality of Opportunity (EOM) metric. $\mathcal{S}$tride-Net consistently occupies the favorable upper-right region, indicating improved fairness without sacrificing predictive accuracy, especially for underrepresented demographic groups. \textcolor{gray}{(Please Zoom In)}.}
% \label{fig:stride_results_ieee}
% \end{figure}

\begin{figure}[t]
    \centering
    \includegraphics[width=\linewidth]{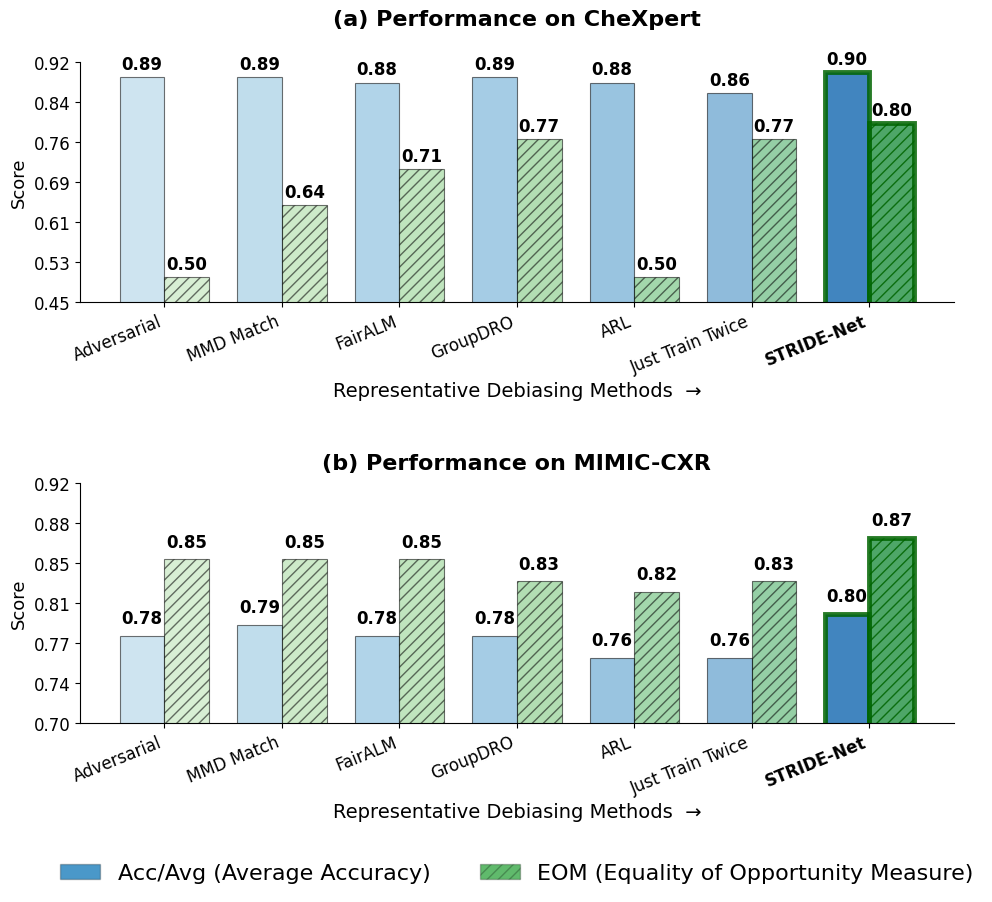}
    \caption{\textbf{Accuracy-fairness trade-offs on CheXpert and MIMIC-CXR.}
    Comparison of $\mathcal{S}$tride-Net with representative debiasing methods under the Equality of Opportunity (EOM) metric. $\mathcal{S}$tride-Net consistently occupies the favorable upper-right region, indicating improved fairness without sacrificing predictive accuracy, especially for underrepresented demographic groups. \textcolor{gray}{(Please Zoom In)}.}
    \label{fig:stride_results_ieee}
    \vspace{-0.3cm}
\end{figure}

\textbf{Intersectional Fairness Analysis}
Fairness challenges are often most severe for intersectional subgroups. When evaluated across race-gender intersections, $\mathcal{S}$tride-Net continues to outperform competing approaches. Accuracy increases to 0.806 on MIMIC-CXR and 0.909 on CheXpert, while both PQD and EOM improve relative to prior methods.

These gains are particularly notable because intersectional groups often suffer from compounding biases. The results suggest that explicitly disentangling disease-relevant features from sensitive attributes at the representation level; as enforced by the GOT alignment and adversarial confusion losses; helps stabilize performance even in low-support subgroups.
\vspace{0.2cm}

\textbf{Accuracy-Fairness Trade-off}
Table~\ref{tab:combined_fairness_results_ieee} reveals distinct trade-off patterns across methods. UBAIA achieves strong PQD but comparatively lower accuracy and EOM, indicating that it equalizes subgroup performance partly by reducing true positive rates for well-performing groups. CheXclusion improves certain fairness metrics but lags behind in overall accuracy.

In contrast, $\mathcal{S}$tride-Net consistently achieves the best balance between accuracy and fairness. By constraining the latent space to contain only label-aligned, stride-selected patches, the model avoids the typical trade-off where fairness improvements degrade clinical utility.
\vspace{0.2cm}

\textbf{Comparison with Debiasing Baselines}
Figure~\ref{fig:stride_results_ieee} situates $\mathcal{S}$tride-Net within the broader landscape of debiasing approaches, including adversarial training~\cite{wadsworth2018achieving}, MMD Match~\cite{pfohl2021empirical}, FairALM~\cite{verma2018fairness}, GroupDRO~\cite{sagawa2019distributionally}, ARL~\cite{lahoti2020fairness}, and Just Train Twice~\cite{liu2021just}. While several baselines achieve modest fairness improvements, these gains often come at the expense of accuracy or exhibit instability across datasets.

$\mathcal{S}$tride-Net consistently occupies the upper-right region of the accuracy-EOM plane, indicating improved fairness without sacrificing predictive performance. This behavior reflects the complementary nature of stride-based localization, semantic alignment, and adversarial disentanglement.
\vspace{0.2cm}

\begin{table}[t]
\centering
\footnotesize
\setlength{\tabcolsep}{5pt}
\renewcommand{\arraystretch}{1.15}
\caption{{Component-wise ablation on ViT-based models.}
Impact of semantic alignment (GOT) and stride-based disentanglement on accuracy and fairness across MIMIC-CXR and CheXpert. 
% Adding GOT improves subgroup stability, while the full $\mathcal{S}$tride-Net yields the strongest gains in both accuracy and fairness, particularly for intersectional subgroups.
}
\label{tab:vit_ablation_combined_ieee}

\begin{tabular}{lccc|ccc}
\toprule
\multirow{2}{*}{Methods $\downarrow$} &
\multicolumn{3}{c|}{MIMIC-CXR} &
\multicolumn{3}{c}{CheXpert} \\
\cmidrule(lr){2-4}\cmidrule(lr){5-7}
& Acc/Avg & PQD & EOM & Acc/Avg & PQD & EOM \\
\midrule

\multicolumn{7}{l}{\textcolor{gray}{Individual Subgroup: ``Race"}} \\
\midrule
ViT-B &
0.772 & \cellcolor{softgreen3}{0.943} & 0.767 &
0.853 & \cellcolor{softgreen3}{0.938} & 0.751 \\

ViT-B+GOT &
0.784 & 0.929 & 0.779 &
0.865 & 0.931 & 0.763 \\

 $\mathcal{S}$tride-Net (Ours) &
\cellcolor{softgreen3}{0.805} & \cellcolor{softgreen1}0.922 & \cellcolor{softgreen3}{0.870} &
\cellcolor{softgreen3}{0.875} & \cellcolor{softgreen1}0.927 & \cellcolor{softgreen3}{0.803} \\

\midrule
\multicolumn{7}{l}{\textcolor{gray}{Intersection Subgroup: ``Race-Gender"}} \\
\midrule
ViT-B &
0.780 & 0.761 & 0.377 &
0.889 & 0.837 & 0.433 \\

ViT-B+GOT &
0.795 & 0.733 & 0.389 &
0.892 & 0.837 & 0.435 \\

 $\mathcal{S}$tride-Net (Ours) &
\cellcolor{softgreen3}{0.806} & \cellcolor{softgreen3}{0.879} & \cellcolor{softgreen3}{0.705} &
\cellcolor{softgreen3}{0.909} & \cellcolor{softgreen3}{0.897} & \cellcolor{softgreen3}{0.466} \\
\bottomrule
\multicolumn{7}{l}{\tiny{High highlight intensity denote high performance}}
\vspace{-0.3cm}
\end{tabular}
\end{table}

\textbf{Component-wise Ablation Study}
To better understand the contribution of each architectural component, we conduct ablation studies using a ViT-B backbone. As shown in Table~\ref{tab:vit_ablation_combined_ieee}, the baseline ViT-B (ERM) achieves strong accuracy but exhibits pronounced fairness gaps, particularly for intersectional subgroups.

Adding the GOT alignment module improves both accuracy and fairness, confirming that aligning patch embeddings with label semantics stabilizes subgroup performance. The full $\mathcal{S}$tride-Net; incorporating the learnable stride mask and adversarial confusion loss; yields the best EOM and substantial PQD improvements while further boosting accuracy.
\vspace{0.2cm}

\textbf{Hyperparameter Sensitivity}
Hyperparameter sweeps reveal that $\lambda_{\mathrm{GOT}} = 0.8$ provides a favorable balance between semantic alignment and task performance. Larger values over-regularize the representation and degrade accuracy. Similarly, setting the confusion loss weight to $\alpha_{\mathrm{conf}} = 2$ yields strong fairness gains with minimal impact on utility, underscoring the importance of moderation in adversarial disentanglement.
% \vspace{0.2cm}

% \subsection*{\textbf{Summary}}

% Overall, the experimental results demonstrate that $\mathcal{S}$tride-Net effectively narrows subgroup performance gaps while maintaining strong predictive accuracy. By enforcing label-aligned, patch-level representations and explicitly discouraging sensitive attribute leakage, the proposed framework delivers a robust and principled approach to fair chest X-ray classification.

\section{Conclusion}
 $\mathcal{S}$tride-Net offers a fairness-aware, disentangled representation learning framework for chest X-ray classification that combines adaptive masking, confusion regularization, and GOT-based semantic alignment. On MIMIC-CXR and CheXpert,  $\mathcal{S}$tride-Net consistently improves fairness across race and race-gender subgroups while preserving or improving diagnostic accuracy. Although this study focuses on chest X-rays and the ``No Finding’’ task, the principles are applicable to other modalities and label sets where sensitive information can be encoded in visual features. Future work will target computational efficiency, interpretability of learned masks and validation in larger, more heterogeneous cohorts.
 % \vspace{-0.2cm}

\bibliographystyle{IEEEtran}
\bibliography{main}

\end{document}